\documentclass[times, preprint, 10pt]{elsarticle}

\usepackage{lineno}

\biboptions{sort&compress}

\usepackage{amssymb,amsmath}
\usepackage[table,xcdraw]{xcolor}
\usepackage{graphicx}
\usepackage{subcaption}
\usepackage{algorithm}
\usepackage{algpseudocode}
\usepackage[normalem]{ulem}
\useunder{\uline}{\ul}{}
\usepackage{pifont}% http://ctan.org/pkg/pifont
\usepackage{hyperref}
\newcommand{\cmark}{\ding{51}}%
\newcommand{\xmark}{\ding{55}}%

\journal{Pattern Recognition}

\begin{document}

\begin{frontmatter}

\title{VISTA: Unsupervised 2D Temporal Dependency Representations for Time Series Anomaly Detection} %% Article title

% a) Make sure your title is succinct and grammatical. It should ideally not exceed 10-15 words.
% b) Make sure your conclusions reflect on the strengths and weaknesses of your work, how others in the field can benefit from it and thoroughly discus future work. The conclusions should be different in content from the abstract, and be rather longer too.
% c) Take a careful look at your bibliography and how you cite papers listed in it. Make sure it is current, and cites recent work. Please cite a variety of different sources of literature. Please do not make excessive citation to arXiv papers, or papers from a single conference series. Do not cite large groups of papers without individually commenting on them. So we discourage " In prior work [1,2,3,4,5,6] …". Your bibliography should only exceptionally exceed about 40 items.
% d) You may have originally written your paper with a different audience in mind. Please make sure the revised version is relevant to the readership of Pattern Recognition. To this end please make sure you cite RECENT  work from the field of pattern recognition that will be relevant to our readership.
% e) Do not exceed the page limits or violate the format, i.e. double spaced SINGLE column with a maximum of 35 pages for a regular paper and 40 pages for a review.

\author[label1]{Sinchee Chin} %% Author name
\author[label1]{Fan Zhang} %% Author name
\author[label2]{Xiaochen Yang} %% Author name
\author[label3]{Jing-Hao Xue} %% Author name
\author[label1]{Wenming Yang\corref{cor1}}%% Author name with corresponding author
\author[label4]{Peng Jia} %% Author name
\author[label5]{Guijin Wang} %% Author name
\author[label6]{Luo Yingqun} %% Author name

\cortext[cor1]{Corresponding author: Shenzhen International Graduate School, Tsinghua University, China. E-mail: yangelwm@163.com}

\affiliation[label1]{organization={Shenzhen International Graduate School, Tsinghua University},%Department and Organization
            city={Shenzhen},
            postcode={518071}, 
            country={China}}

\affiliation[label2]{organization={University of Glasgow},%Department and Organization
            city={Glasgow}, 
            postcode={G12 8QQ}, 
            country={United Kingdom}}

\affiliation[label3]{organization={University College London},%Department and Organization
            city={London}, 
            postcode={WC1E 6BT}, 
            country={United Kingdom}}

\affiliation[label4]{organization={Collaborative Innovation Center for Transport Studies, Dalian Maritime University},%Department and Organization
            city={Dalian}, 
            postcode={116028}, 
            country={China}}

\affiliation[label5]{organization={Department of Electric Engineering, Tsinghua University, },%Department and Organization
            city={Beijing}, 
            postcode={100190}, 
            country={China}}

\affiliation[label6]{organization={Shenzhen Skyworth AI Co. ,Ltd},%Department and Organization
            city={Shenzhen}, 
            postcode={518027}, 
            country={China}}

%% Abstract
\begin{abstract}
Time Series Anomaly Detection (TSAD) is essential for uncovering rare and potentially harmful events in unlabeled time series data. 
However, existing methods are highly dependent on clean, high-quality inputs to perform effectively, making them susceptible to noise and real-world imperfections. 
Additionally, the intricate temporal relationships in time series data are often inadequately captured in traditional 1D representations, leading to suboptimal modeling of temporal dependencies.
We introduce \textbf{VISTA}, a training-free, unsupervised TSAD algorithm designed to overcome these challenges. VISTA features three core modules: 
\textbf{(1) Time Series Decomposition}, which uses Seasonal and Trend Decomposition via Loess (STL) to decompose noisy time series into trend, seasonal, and residual components and leverage them to represent the complex patterns embedded in time series; 
\textbf{(2) Temporal Self-Attention}, which transforms one-dimensional time series into two-dimensional temporal correlation matrices, enabling richer modeling of temporal dependencies for visualization and anomaly detection of time discrepancies; 
and \textbf{(3) Multivariate Temporal Aggregation}, which leverages a pretrained feature extractor to integrate information across all variables into a unified, compact, and memory-efficient representation. 
VISTA's training-free approach allows for rapid deployment and easy hyperparameter tuning, making it suitable for industrial applications. 
VISTA achieves state-of-the-art performance on five commonly used multivariate TSAD datasets.
The code will be released \href{https://github.com/SinChee/VISTA}{here}.
\end{abstract}

\begin{highlights}
    \item Introduce VISTA, a training-free TSAD algorithm modeling 2D temporal dependencies.
    \item Propose a 2D transformation strategy for visualization using pretrained CNNs in TSAD.
    \item Achieve state-of-the-art performance on five multivariate time series datasets.
\end{highlights}


\end{frontmatter}

% \linenumbers

\section{Introduction}
\label{sec:introduction}
The rapid advancement of technology, particularly with the proliferation of wireless communication and Internet of Things (IoT) devices, has led to an explosion of time series data across various industries. 
However, the increasing volume and complexity of time series data have amplified the challenges of analyzing and detecting anomalies, especially when multiple sensors are involved.
Furthermore, identifying such anomalies is particularly challenging due to their rarity and context dependence, often requiring specialized methods to differentiate them from normal variations.
Recently, time series anomaly detection (TSAD) has emerged as a critical task for detecting these complex multivariate anomalies, leveraging temporal dependencies and patterns to identify deviations that may indicate faults, irregularities, or rare events in dynamic systems, such as system failures \cite{chen2021thermodynamic}, financial fraud \cite{acien2022becaptcha}, or medical emergencies \cite{dissanayake2023multi}. 

TSAD faces several critical challenges that hinder its effectiveness in real-world applications.
First, time series data are noisy, which poses significant challenges for anomaly detection models, especially reconstruction-based methods \cite{wu2023timesnet}. Noise often obscures meaningful patterns, making it difficult for models to generalize and detect anomalies accurately. 
While some methods attempt to denoise the input time series, they risk inadvertently removing anomalies, as certain anomalies manifest as noise \citep{sui2024deep}. 
This trade-off between preserving anomalies and reducing noise adds further complexity to the modeling process, especially in real-world scenarios where the distinction between noise and anomalies is often subtle.
Second, the high dimensionality of multivariate time series data significantly increases its complexity, especially in scenarios involving codependent anomalies \cite{zamanzadeh2024deep}. 
The distinct states represented by different variables create highly non-linear decision boundaries, making it particularly challenging to differentiate between normal and anomalous conditions.

A common approach to TSAD is to analyze the temporal dependencies. 
Understanding the temporal dependencies of time series data helps in understanding the temporal and contextual factors. 
For example, in industrial machinery monitoring, vibration levels may differ significantly based on the machine’s operational mode (indicated by the input power). During high-speed operation, vibration levels may consistently fall within a specific range, while during low-speed or idle states, much lower levels may still be considered normal \cite{wang2025fault}. 
To address temporal anomalies, various methods have been proposed to capture temporal dependencies.
Reconstruction-based methods leverage models such as transformers \cite{tuli2022tranad}, autoencoders \cite{park2018multimodal}, and diffusion models \cite{xiao2023imputation}. 
These methods aim to encode the input, denoise anomaly points during the encoding-decoding phase, and reconstruct the time series. 
Anomalies are identified when the reconstructed value deviates significantly from the input signal. 
TimesNet~\cite{wu2023timesnet} proposes a 1D-to-2D transformation by analyzing the dominant frequency of the time series.
Although they show promising results in representing the temporal dependencies of 1D time series in 2D, they still suffer from noisy training data, which can lead to overfitting and inefficient learning \cite{si2024timeseriesbench}. 
Moreover, the presence of noise in training samples poses a significant challenge, making it difficult to distinguish true anomalies from noise and limiting the effectiveness of these methods in real-world scenarios.
Another approach to TSAD involves leveraging latent representations of time series data for anomaly detection, often using contrastive learning to distinguish between normal and abnormal samples \cite{zhang2022self}. 
While effective in certain cases, these methods also struggle to handle noisy time series inputs, where noise can obscure patterns critical for anomaly detection. 
Furthermore, they rely on prior assumptions about anomaly distributions, which may not generalize well across diverse scenarios. 
This highlights a fundamental challenge for TSAD: 

\textit{How can we detect anomalies in time series data corrupted by noise with an unknown distribution?}

To address the challenges inherent in TSAD, we propose VISTA, a novel, representation-based, training-free TSAD algorithm comprising three core modules: \textbf{Time Series Decomposition}, \textbf{Temporal Self-Attention}, and \textbf{Multivariate Temporal Aggregation}. Each module is meticulously designed to mitigate noise, enhance interpretability, and integrate information from multivariate time series data.
VISTA begins with the decomposition of noisy time series data into three distinct components—trend, seasonal, and residual—using Seasonal and Trend decomposition via Loess (STL) \cite{cleveland1990stl}. The trend component captures the overall movement of the data, effectively smoothing noise; the seasonal component identifies periodic patterns, incorporating cyclical information; and the residual component isolates discrepancies and unpredictable variations, providing a crucial representation of both anomalies and noise.
Subsequently, the Temporal Self-Attention module integrates these components into a Temporal Correlation Matrix, employing self-attention mechanisms to model temporal relationships. This matrix, represented as an interpretable RGB image, serves as input to a pretrained Convolutional Neural Network (CNN) for feature extraction. CNNs' robustness to noise and their ability to exploit local spatial context \footnote{Temporal context in this application} ensure reliable and effective feature extraction.
For multivariate time series, VISTA aggregates variables by summing their representations, ensuring memory efficiency, and constructing a compact yet comprehensive representation. This approach maintains scalability while preserving the robustness necessary for multivariate anomaly detection tasks.
Overall, this paper presents the following contributions:
\begin{itemize}
    \item We propose VISTA (\textbf{V}isualizing \textbf{I}nsight\textbf{S} with \textbf{T}emporal Dependency in Time-series \textbf{A}nomalies), a training-free Time Series Anomaly Detection algorithm that models temporal dependencies, enabling rapid deployment and reducing computational overhead.
    \item We introduce a 2D temporal dependency visualization method that preserves both signal and noise, effectively representing temporal components using a pretrained CNN. This is the first training-free 2D transformation leveraging temporal dependency matrices for both visualization and analysis, offering interpretability and robustness.
    \item VISTA achieves overall state-of-the-art performance on five multivariate time series datasets, outperforming baselines in Standard F1 Score and ROC-AUC, demonstrating its effectiveness and generalizability.
\end{itemize}

\section{Related Work}
\label{sec:related-work}

Early approaches relied on statistical methods, leveraging assumptions about specific patterns in time series data to enable anomaly detection \cite{domingues2018comparative}. These methods provided interpretable solutions for identifying anomalies in cyclical and linear patterns.
Furthermore, classical machine learning algorithms, such as Principle Clustering Analysis (PCA) \cite{scholkopf2001estimating} and Isolation Forest \cite{liu2008isolation} have been used to detect time series anomalies.
Statistical methods offer valuable insights into the temporal structure of time series data. 
They excel at identifying stationary patterns and cyclical behaviors, with Autoregressive Integrated Moving Average Model (ARIMA) \cite{box1970distribution} enhancing the ability to model non-stationary data through differencing operations.
Classical machine learning approaches, on the other hand, introduced a data-driven perspective, enabling the modeling of nonlinear relationships and interdependencies between features.
However, these approaches exhibit significant limitations. 
Statistical methods struggle with high dimensionality and intricate temporal dependencies, making them less effective for multivariate time series. 
Classical machine learning models often depend heavily on feature engineering, limiting their scalability and adaptability to diverse datasets. 
Additionally, while some methods are effective in handling noise, distinguishing true anomalies from noisy variations remains a persistent challenge, reducing their robustness in real-world applications.
Recent research underscores the value of revisiting and integrating classical methods into deep learning anomaly detection frameworks for multivariate time series \cite{audibert2022deep}.

% 2. Deep learning methods
Benefiting from the strong nonlinear modeling and feature extraction capabilities of deep neural networks, deep learning methods have become the mainstream in current research of TSAD. Broadly, these methods can be categorized into three types: (i) Forecasting-based methods, which predict a point or recent window and use the deviation between the forecasted and actual values as an anomaly score. (ii) Reconstruction-based methods, which reconstruct the time series data to better capture the underlying patterns and detect anomalies \cite{lee2024explainable}. (iii) Representation-based methods, which map time series data into a feature space and leverage latent representations for anomaly detection \cite{zong2018deep}.
Forecasting and reconstruction methods have garnered significant attention due to their superior performance. 
To incorporate temporal information efficiently, a Temporal Convolutional Network (TCN) is used to predict the subsequence time series and detect time series anomalies \cite{tian2024mfgcn}.
Recent work utilizes transformers to capture time dependencies for time series data, leveraging a self-attention mechanism to reconstruct input time series for time series anomaly detection \cite{xu2022anomaly}. 
IGCL \cite{zhao2024unsupervised} and MultiRC \cite{hu2024multirc} predict anomalies through the indication of an anomaly precursor. 
However, the anomaly precursor is ambiguous, resulting in poor anomaly detection.
Although these methods show promising results in TSAD, they do not address the inherent noise present in the training data.
This greatly hinders the ability of deep models to learn a good reconstruction and is prone to overfitting.
Additionally, the successful application of contrastive learning in time series anomaly detection has led to highly competitive detection performance for representation-based methods. DCdetector \cite{yang2023dcdetector} integrates deep CNNs and dual attention mechanisms to focus on spatial and temporal dimensions, enhancing the separability of normal and anomalous patterns through contrastive learning. CARLA \cite{darban2025carla} employs a two-stage framework: it first distinguishes between anomaly-injected samples and original samples and then uses neighborhood information for self-supervised classification to boost anomaly detection. 
Notably, CARLA trains data from different entities within each dataset separately, deviating from the common practice of training all data together.

Recently, researchers leverage the two-dimensional representation of time series data for TSAD. TimesNet \cite{wu2023timesnet} decomposes time series into multiple characteristic periods using fast Fourier transforms and establishes two-dimensional representations through stacked layers, leveraging reconstruction errors for anomaly detection. 
% ITF-TAD \cite{namura2024training} applies wavelet transforms to convert time series data into spectrograms, compressing them into single representations.
Young-Joo Hyun et al. \cite{hyun2024encoding} transformed time series into the frequency domain and applied recurrence plot methods to convert them into images. 
However, these frequency-domain approaches lose temporal information and are unable to achieve point-based anomaly detection.

Unlike previous 2D representation methods, VISTA encodes temporal dependencies into a 2D Temporal Correlation Matrix, consisting of three distinct channels that represent the trend, seasonal, and residual components. This approach ensures that no essential information is removed from the raw input signal. The trend channel effectively reduces noise, the seasonal channel accurately captures periodic transformations, and the residual channel preserves critical anomalies and deviations. Each 2D Temporal Correlation Matrix is designed to enable seamless extraction of features using pretrained CNNs. 

\section{Methodology}

% Use figure* for multi-column figure
\begin{figure*}[tp]
    \centering
    \includegraphics[width=\linewidth]{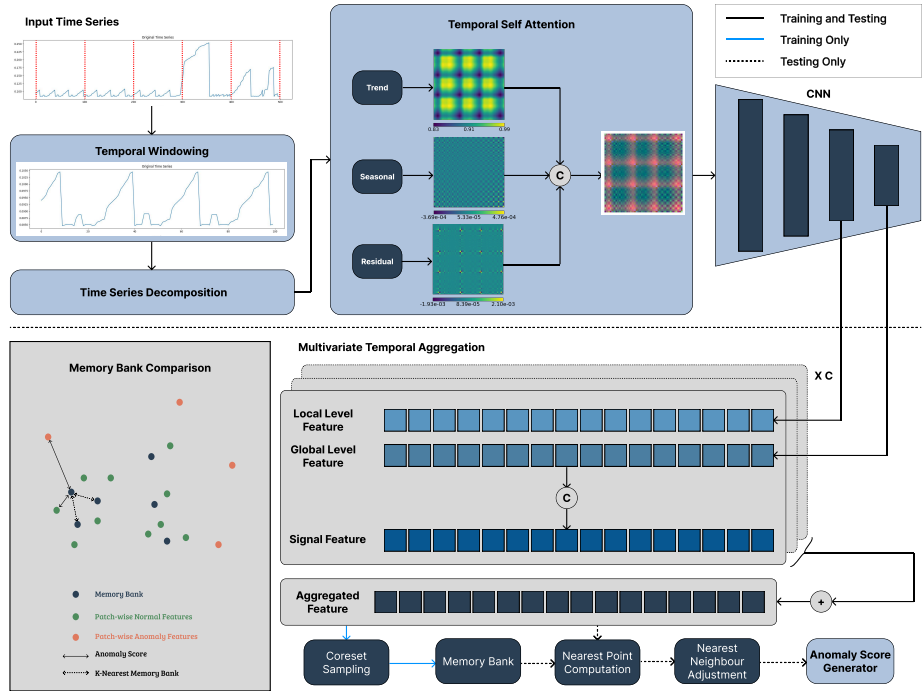}
    \caption{
    Overall architecture of VISTA. The proposed architecture processes long time series data in several stages. Initially, the data are partitioned into fixed-size windows of length $w_s$. Each window is then decomposed using STL decomposition into trend, seasonal, and residual components, each capturing distinct and significant aspects of the time series. 
    The Temporal Self-Attention Module computes a temporal correlation matrix for each time series variables, enhancing interpretability of temporal dependencyfor each decomposed components. The temporal correlation matrices are further processed using a pretrained CNN network to extract intermediate features. 
    Multivariate Temporal Aggregation Module is employed to aggregate features across each time series variable, yielding a unified feature representation. A greedy sampling algorithm is then applied to select the most representative features, which are stored in a memory bank for efficient retrieval. 
    During inference, the memory bank constructed during training is used to compare query features via $L_2$-distance. To further enhance detection accuracy, the anomaly scores are rescaled using the method proposed in \cite{roth2022towards}.
    }
    \label{fig:overall}
\end{figure*}

\label{sec:methodology}

\textbf{Preliminaries:} We consider a collection of multivariate time series denoted by $\mathcal{X}$, which contains signal recorded from timestamp 1 to \( T \). Specifically:
\begin{equation}
    \mathcal{X} = \{\mathbf{x}_{1},\mathbf{x}_{2},\cdots,\mathbf{x}_{T}\},
\end{equation}
where \(\mathbf{x}_{t}\in \mathbb{R}^{C} \) represents an \(C\)-dimensional vector at timestamp \( t \). 
The goal of the multivariate time series anomaly detection task is to determine whether the observation \(\mathbf{x}_{t} \) is anomalous. 
The state at timestamp \( t \) is represented by \( \mathbf{y}_{t} \in \{0,1\} \), where 0 indicates normal and 1 indicates anomalous. 
We define a CNN with ImageNet-pretrained weights as \(f_\theta(\cdot)\) and the intermediate feature maps at the \(i\)-th layer be \(F_{i} \in \mathbb{R}^{H_i \times W_i \times L_i}\), where \(H_i\), \(W_i\), and \(L_i\) are the height, width, and channel number of the \(i\)-th layer. 

\subsection{Overview}
\label{sec:method-overall}
In this section, we present the proposed VISTA (\textbf{V}isualizing \textbf{I}nsight\textbf{S} with \textbf{T}emporal Dependency in Time-series \textbf{A}nomalies), a representation-based TSAD for multivariate time series data.
VISTA is a training-free TSAD algorithm that leverages STL decomposition to decouple time series components, suppressing irrelevant noise. 
Next, VISTA models temporal dependencies leveraging the autocorrelation of trend, seasonal, and residual signals decomposed from the original time series. 
Finally, we employ a pre-trained CNN to extract the local features for normal temporal correlation modeling. 
% VISTA is a training-free TSAD algorithm, leveraging STL decomposition, temporal dependency self-attention, and pretrained CNN to model the normal training samples. 
The training-free nature of VISTA offers two significant advantages. 
First, it enables easy deployment and rapid parameter tuning, which is crucial for timely anomaly detection in dynamic environments. 
Second, it demonstrates the efficacy of our temporal correlation matrix approach by showing that pretrained ImageNet CNNs can effectively distinguish between normal and anomalous patterns without additional training.
This approach leverages the rich feature representations learned from large-scale image datasets, ensuring robust anomaly detection capabilities.
We present the overall methodology of VISTA in Figure~\ref{fig:overall}.

\subsection{Time Series Decomposition}
% we use stl to decompose into trend, seasonal, and residual
% show some equation for this decomposition
The first step of VISTA is to segment the raw time series data into non-overlapping windows of fixed size \( w_s \), resulting in \( \mathbf{X} = \{\mathbf{X}_1, \mathbf{X}_2, \dots, \mathbf{X}_n\} \), where each window \( \mathbf{X}_i \in \mathbb{R}^{w_s \times C} \). 
Next, we decompose each time series window independently using \textit{Seasonal and Trend Decomposition using Loess (STL)} into three components: \textbf{trend}, \( \mathbf{T}_{i} \in \mathbb{R}^{w_s \times C} \); \textbf{seasonal}, \( \mathbf{S}_{i} \in \mathbb{R}^{w_s \times C} \); and \textbf{residual}, \( \mathbf{R}_{i} \in \mathbb{R}^{w_s \times C} \). 
Mathematically, the original time series can be expressed as the summation of these three components as  
\begin{equation}
    \mathbf{X}_i = \mathbf{T}_{i} + \mathbf{S}_{i} + \mathbf{R}_{i}.
    \label{eq:decompose}
\end{equation}
To ensure consistent STL decomposition across different temporal windows, we set the seasonal period to 50\% of the window size. This approach strikes a balance between capturing low-frequency (long-term) trends and high-frequency (short-term) variations, avoiding over-smoothing that can occur with larger seasonal periods while preserving finer periodic patterns. Experimental analysis shows that this choice has minimal impact on performance, as the decomposition components dynamically adjust to the underlying structure of the time series.

% trend, seasonal and residual meaning in this context
The trend component captures the underlying long-term progression of the time series, such as gradual increases, decreases, or persistent shifts over time. 
This component eliminates high-frequency noise by smoothing the data, providing a stable baseline for detecting significant deviations. 
The seasonal component represents periodic patterns or repeating behaviors inherent in the time series.
Isolating this component enhances the temporal periodicity in the data, helping to distinguish between normal cyclical behavior and anomalous variations. 
The residual component encapsulates noise and any deviations that cannot be expressed by the trend or seasonal components. 
It serves as a critical focus for anomaly detection, as anomalies are often hidden within these unexplained variations. 
By explicitly modeling residuals, we amplify their representation, making irregularities easier to identify compared to analyzing raw time series data.
This decomposition forms the foundation of our methodology, facilitating TSAD by suppressing noise, incorporating periodicity learning, and enhancing anomaly deviation.
For simplicity, we omit the index for each window, and the subsequent equations are indexed with \(c\) representing each variable in the multivariate time series.

\subsection{Temporal Self-Attention}
% how TSA works 
Following the decomposition step, we employ a Temporal Self-Attention Module to transform each decomposed component from a one-dimensional (1D) sequence into a two-dimensional (2D) representation, facilitating the visualization of temporal dependency within each fixed-size window.
The temporal self-attention mechanism operates by computing a temporal correlation matrix for each decomposed component and each variable of multivariate time series. 
The \textbf{temporal correlation matrix} \( \mathbf{A}_c \in \mathbb{R}^{w_s \times w_s \times 3} \) is defined as 
% \begin{equation}
%     \mathbf{A}_c = \lambda \left( \hat{\mathbf{X}}_c \cdot \hat{\mathbf{X}}_c^\top \right)
%     \label{eq:sa-matrixing}
% \end{equation}
\begin{equation}
    \mathbf{A}_c = {\rm concat}(\mathbf{T}_c \times \mathbf{T}_c^\top, \ \mathbf{S}_c \times \mathbf{S}_c^\top, \ \mathbf{R}_c \times \mathbf{R}_c^\top),
    \label{eq:sa-matrixing}
\end{equation}
% Where \( \lambda \) is a scaling factor that modulates the intensity of the correlation scores. 
which encapsulates the self-similarity between each pair of time points within the window, efficiently capturing the time dependency.
It is crucial to note that we do not perform normalization to \( \mathbf{A}_c \) because we want to maintain the magnitude of each component, effectively capturing temporal relationships, seasonal pattern change, and noise deviation.

% why we do temporal 2D transformation
% first it helps in anomaly analysis
The introduction of the temporal correlation matrix serves dual purposes in time series anomaly detection. Firstly, it provides a visual representation of temporal correlations, aiding in the analysis and interpretation of complex time series data. Figure~\ref{fig:tsa-vis} highlights the stark differences in temporal correlation matrices between normal and anomalous windows, demonstrating the module’s effectiveness in capturing disruptions in temporal patterns.
For a constant signal, the seasonal and residual components have negligible effects on the corresponding temporal correlation matrix, leaving it primarily influenced by the trend component. An anomaly in the constant signal creates a significant change in the temporal correlation matrix, where a distinct red dot emerges at the location of the anomaly, surrounded by a blue circle, indicating localized disruptions.
In the case of a periodic signal, the trend component typically exhibits consistent distances between peaks, while the seasonal component forms a grid-like structure. A seasonal change in the periodic signal stretches the distances between peaks and distorts the grid pattern, highlighting the disruption in periodicity.
For an ascending signal, the temporal correlation matrix reveals a steady increment from the top-left to the bottom-right, corresponding to the increasing trend. When there is a sudden deviation in the incremental trend, the matrix distinctly captures the discontinuity, making anomalies visually apparent.
The visual nature of temporal correlation matrices facilitates intuitive analysis of temporal dependencies, enabling feature extractors to leverage these patterns for robust and effective anomaly detection. This enhances the model’s ability to identify subtle temporal discrepancies that may be challenging to detect with traditional methods.

Next, the 1D-to-2D transformation ensures the compatibility with the input to CNN, enabling us to leverage powerful pretrained CNNs for feature extraction. 
CNNs are inherently more robust to noise compared to models directly processing raw time series inputs due to their architectural design and operational principles \cite{goodfellow2014explaining}. 
Therefore, the temporal correlation matrix not only serves as a powerful visualization for time dependencies,
it also enables the use of CNN to extract distinguishable features for modeling and anomaly detection.
Previous methods \cite{wu2023timesnet} also leverage CNN architecture for feature extraction from noise inputs. 
However, the input is extremely noisy or under-represented, which greatly hinders effective feature learning. 
The temporal self-attention module not only preserves the original information from the training samples, 
it also helps in enhancing the effect of seasonal and noise patterns.

\subsection{Multivariate Temporal Aggregation}
\label{sec:method-multi-channel-aggregation}

Multivariate time series data capture signals from multiple sensors, with each variable providing distinct information specific to its respective sensor. Anomalies in such data often manifest in a codependent manner, where interactions between variables play a critical role in identifying abnormal patterns \cite{zamanzadeh2024deep}. To address this, we utilize a signal-wise feature extraction and summation approach, ensuring that the unique representations of individual variables are preserved while effectively integrating information across variables.

\paragraph{Feature Extraction}
Each temporal correlation matrix is downsampled to \(32 \times 32\) to ensure memory efficiency for time series anomaly detection.
Next, we employ a {signal-wise feature extraction} strategy, ensuring independent feature extraction from different variables. 
We extract \(l^{\text{th}}\)-intermediate layers' feature, \(\mathbf{F}_{c}^{(l)} \in \mathbb{R}^{H_l \times W_l \times L_l}\) from the CNN model and concatenate them to form the signal feature map, \( \mathbf{F}_{c}^{\text{sig}} \). 
Practically, we choose the third and fourth layers because we require a deep semantic information for detecting time series anomalies. Equation \ref{eq:feature-signal} shows the computation of the signal feature map.

\begin{equation}
   \mathbf{F}_{c}^{\text{sig}} = \text{concat} \left( \mathbf{F}_{c}^{(3)}, \text{interpolate}(\mathbf{F}_{c}^{(4)}, (H_3, W_3)) \right)
   \label{eq:feature-signal}
\end{equation}

\paragraph{Feature Aggregation}
After feature extraction, the signal feature maps from different variables are aggregated through simple summation. Specifically, the signal feature maps from each variable \( c \) are summed to form the Aggregated Feature Map, \(\mathbf{F}_{\text{agg}} \in \mathbb{R}^{H_3 \times W_3 \times (L_3 + L_4)}\)  as shown in Equation \ref{eq:aggregated_feature}.
\begin{equation}
    \mathbf{F}_{\text{agg}} = \sum_{c=1}^{C} \mathbf{F}_{c}^{\text{sig}}.
    \label{eq:aggregated_feature}
\end{equation}
This summation performs the final aggregation from the multivariate time series data into a unified, memory-efficient, and compact representation. We then use each patch from \(\mathbf{F}_{\text{agg}} \) to build the memory bank for anomaly detection.

\paragraph{Memory Bank Construction}
The \textbf{memory bank} $\mathcal{M}$ is constructed to serve as a compact repository of representative feature vectors that encapsulate all patches from the aggregated features. Let $\mathcal{M}_{j} = \{\mathbf{m}_1, \mathbf{m}_2, \dots, \mathbf{m}_P\}$ represent the set of feature vectors corresponding to the \(i^\textbf{th}\) aggregated feature, \(\mathbf{F}_{\text{agg}, i} \), where \(\mathbf{m}_p \in \mathbb{R}^{L_3 + L_4}\) is the feature vector for the \(p\)-th patch, and \(P= H_3 \times W_3\). Then, the overall memory bank $\mathcal{M}$ is defined as the union of all $\mathcal{M}_j$ from $j = 1$ to $j = n$:

\[
\mathcal{M} = \bigcup_{j=1}^{n} \mathcal{M}_{j}.
\]

To reduce redundancy and prevent overfitting, we employ a \textit{greedy coreset selection} strategy~\cite{roth2022towards} to subsample the memory bank while retaining its coverage of the feature space. The coreset selection from \(\mathcal{M}\) optimizes the following objective:
\begin{equation}
    \mathcal{M}_B = \underset{\mathcal{M}_B \subseteq \mathcal{M}}{\mathrm{argmin}} \max_{\mathbf{n} \in \mathcal{M}} \min_{\mathbf{m} \in \mathcal{M}_B} \|\mathbf{m} - \mathbf{n}\|_2,
    \label{eq:greedy}
\end{equation}
where \(\mathcal{M}_B\) is the coreset, a reduced subset of \(\mathcal{M}\). 
The problem as shown in Equation \ref{eq:greedy} is NP-hard, the computation of \(\mathcal{M}_B\) is shown in Algorithm \ref{alg:coreset}.
The greedy sampling algorithm only samples \(K_B\) points from \(\mathcal{M}\) and the resulting \(\mathcal{M}_B\) can closely approximate \(\mathcal{M}\) while significantly reducing its size.

\begin{algorithm}[htbp]
    \caption{Greedy Coreset Selection for Memory Bank Construction}
    \label{alg:coreset}
    \begin{algorithmic}[1]
    \Require Set of feature vectors \(\mathcal{M} = \{\mathbf{m}_1, \mathbf{m}_2, \dots, \mathbf{m}_N\}\), target coreset size \(K_B\)
    \Ensure \(|\mathcal{M}| \ge K_B\)
    
    \State \(\mathcal{M}_B \gets \{\mathbf{m}_i\}\), where \(\mathbf{m}_i\) is randomly selected from \(\mathcal{M}\)
    \While{\(|\mathcal{M}_B| < K_B\)}
        \ForAll{\(\mathbf{m} \in \mathcal{M} \setminus \mathcal{M}_B\)}
            \State Compute distance to the closest vector in \(\mathcal{M}_B\):
            \[
            d(\mathbf{m}) = \min_{\mathbf{n} \in \mathcal{M}_B} \|\mathbf{m} - \mathbf{n}\|_2
            \]
        \EndFor
        \State Select \(\mathbf{m}^* = \arg\max_{\mathbf{m} \in \mathcal{M} \setminus \mathcal{M}_B} d(\mathbf{m})\)
        \State Add \(\mathbf{m}^*\) to \(\mathcal{M}_B\): \(\mathcal{M}_B \gets \mathcal{M}_B \cup \{\mathbf{m}^*\}\)
    \EndWhile
    \State \Return \(\mathcal{M}_B\)
    \end{algorithmic}
    \end{algorithm}

\subsection{Inference Stage}
\label{sec:inference}

During the inference stage, the test time temporal correlation matrix, \(\mathbf{A}_c^{\text{test}}\) is passed to the pretrained CNN for feature extraction. 
The resulting test time aggregated features vector, \(\mathbf{f}_p^{\text{test}} \in \mathbb{R}^{L_3+L_4}\) are compared with the nearest point in \(\mathcal{M}_B\), leveraging the Euclidean Distance. 
% \textcolor{red}{(JHX: What is $\mathcal{P}_\text{test}$? What is its relationship wtih $\mathcal{F}_\text{patch}$? Or do you mean $\mathcal{F}_\text{patch}^{\rm test}$?)}
% \textcolor{blue}{(CSC: refine equation and symbols accordingly.)}
The computation of the anomaly score is shown in Equation \ref{eq:anomlay-score}.
\begin{equation}
    s_p^{*}= \min_{\mathbf{m} \in \mathcal{M}_B} \|\mathbf{f}_p^{\text{test}} - \mathbf{m}\|_2,
    \label{eq:anomlay-score}
\end{equation}
where \(s_p^{*}\) represents the anomaly score for the \(p\)-th feature vector in the test time aggregated feature, \(\mathbf{F}_{\text{agg}}^{\text{test}}\). 
The greater the anomaly score, the more likely the patch is an anomaly.
% \textcolor{red}{(JHX: The higher, the more like an anomaly?)}.
% \textcolor{blue}{(CSC: Yes it is, added a explanation to this, but sure if we should say this, because Eq (8) explained it.)}.
To further refine the anomaly score, we leverage the density of the nearest memory point to rescale the anomaly score, ensuring a more robust anomaly detection \cite{roth2022towards}.
% \textcolor{red}{(JHX: Please add citation for this strategy.)} 
% \textcolor{blue}{(CSC: Citation added for this strategy.)}. 
Equation \ref{eq:anomaly-rescale} shows the reweighting function for the anomaly score by considering the \(K\) nearest memory points.
\begin{equation}
    s_{i} = \left(1 - \dfrac{ \text{exp} (\| \mathbf{m^{*}} - \mathbf{m} \|_2)}{\sum_{\mathbf{m}_k\in \mathcal{N}_K(m^{*})} \text{exp} (\| \mathbf{m^{*}} - \mathbf{m}_k \|_2)} \right) \cdot s_i^{*}
    \label{eq:anomaly-rescale}
\end{equation}
where \(\mathbf{m^{*}}\) is the closest memory feature to the query patch and \(\mathcal{N}_K(m^{*})\) is the function that returns the \(K\) nearest points from \(m^{*}\).
% \textcolor{red}{(JHX: What is $\mathbf{m}_k$?)}.
% \textcolor{blue}{(CSC: added description to $\mathbf{m}_k$)}.
For memory points that are scares, it is likely to be corruptions to the signal, using this anomaly score rescaling method, we can increase the anomaly score on these scarce events in the memory bank.

To obtain the original time series window length, we upsample the anomaly map to \(\mathbb{R}^{w_s \times w_s}\).
The upsampled anomaly map derived from the temporal correlation matrix highlights inconsistencies in temporal relationships and provides point-based anomaly detection by highlighting regions where temporal dependency is violated.
% \textcolor{red}{(JHX: Please define what an anomaly map is; define what its column (or say vertical components?) means? And its row (or say horizontal components?))}
% \textcolor{blue}{(CSC: vertical components or horizontal components is the same, we just choose to sum the vertical component, added description to anomaly map.)}
We sum the vertical components of the upsampled anomaly map to compute the time series anomaly score, aggregating discrepancies over different time steps. 
% \textcolor{red}{(JHX: Please give the relationship between the  $S_(t)$ below and the $s_i$ in Eq(10).)}
% \textcolor{blue}{(CSC: added explanation to S)}
Finally, the time series anomaly scores, \(S=\{s(1), s(2), \dots s(T) \}\) are compared against a predefined threshold \(\tau\) to detect anomalies in the time series data. If \(S > \tau\), the input is flagged as anomalous; otherwise, it is classified as normal:

\begin{equation}
    \text{Prediction}(t) =
    \begin{cases}
    \text{1}, & \text{if } s(t) > \tau, \\
    \text{0}, & \text{otherwise}.
    \end{cases}
\end{equation}

\section{Results and Discussion}
\label{sec:results}

\subsection{Experimental Setup}
\paragraph{Datasets}
We benchmark VISTA with five commonly used public datasets for multivariate time series anomaly detection.
(1) Mars Science Laboratory rover (MSL)~\cite{hundman2018detecting} contains operational data from multiple sensors onboard a Mars rover. 
(2) Soil Moisture Active Passive satellite (SMAP)~\cite{hundman2018detecting} provides soil moisture data collected by satellite sensors.
(3) Server Machine Dataset (SMD)~\cite{su2019robust} is a five-week dataset collected from a large internet company.
(4) Pooled Server Metrics (PSM)~\cite{abdulaal2021practical} comprises data collected from internal nodes of multiple application servers at eBay. 
(5) Secure Water Treatment (SWaT)~\cite{mathur2016swat} includes sensor readings and actuator states recorded during the operation of a small-scale real-world water treatment system. All training sets consist only of unlabeled data, while the test sets contain data with anomaly labels. The statistical details of the datasets are summarized in Table \ref{table_data_distribution}.

\paragraph{Baselines}
We evaluate the performance of VISTA by benchmarking it against 11 multivariate time series anomaly detection models. 
Forecasting and reconstruction-based methods, including Anomaly Transformer~\cite{xu2022anomaly}, PAD~\cite{jhin2023precursor}, OmniAnomaly~\cite{su2019robust}, GANomaly~\cite{du2021gan}, CAE-Ensemble~\cite{campos2021unsupervised}, TimesNet~\cite{wu2023timesnet}, D3R~\cite{wang2024drift}, and IGCL~\cite{zhao2024unsupervised}, detect anomalies by modeling normal temporal patterns and identifying deviations during forecasting or reconstruction. 
Representation-based methods, such as Deep SVDD~\cite{ruff2018deep}, DAGMM~\cite{zong2018deep}, and DCDetector~\cite{yang2023dcdetector}, leverage compact latent representations to distinguish between normal and abnormal behaviors. 
The experiments for the baselines are conducted using their recommended hyperparameters as specified in their official release source code.

\paragraph{Implementation Details} 
We implement VISTA using PyTorch 2.5.0 and Anomalib \cite{akcay2022anomalib}, a widely used library for industrial visual anomaly detection.
All experiments are conducted on 1 NVIDIA 3090 GPUs and an AMD EPYC 7443 24-core processor.
As a training-free approach, VISTA achieves optimal performance across different datasets by tuning its hyperparameters. 
We perform a grid search to identify the best configurations, such as selecting the sliding window size from \{32, 64, 128, 256, 512, 1024\}, the coreset sampling ratio from 0.1 to 0.9, and the number of nearest neighbors for anomaly score scaling from 5 to 15. 
ResNet-18 is used as the default feature extractor, with the third and fourth intermediate features chosen for representing the samples.

\begin{table}[t]
\caption{Dataset details, which include the domain of the dataset, the number of variables in the time series data (dimension), the number of samples in the training and test sets, and the anomaly ratio between anomalous data points to the normal data points in the test set.}
\resizebox{\columnwidth}{!}{
\begin{tabular}{cccccc}
\hline
Datasets & Domain          & Dimension & Train Set & Test Set & Anomaly Ratio \\ \hline
MSL       & Spacecraft      & 55        & 58,317    & 73,729   & 10.50\%       \\
SMAP      & Spacecraft      & 25        & 135,183   & 427,617  & 12.80\%       \\
SMD       & Server Machine  & 38        & 708,405   & 708,420  & 4.20\%        \\
PSM       & Server Machine  & 25        & 132,481   & 87,841   & 27.80\%       \\
SWaT      & Water Treatment & 51        & 496,800   & 449,919  & 12.10\%       \\ \hline
\end{tabular}
}
\label{table_data_distribution}
\end{table}

\paragraph{Evaluation Metrics}
% we are not going to use PA
We evaluate the performance of time series anomaly detection methods using four metrics, namely precision, recall, F1 score, and ROC-AUC. 
The threshold for precision, recall, and F1 score is computed from the optimal F1 Score.
The ROC-AUC is a threshold-free evaluation metric with 0.5 indicating random performance. 
Previous work has mostly adopted the point adjustment method, which considers all anomalies in the same segment to be correctly detected if a single point within the segment is correctly identified. 
However, relying on ground truth or prior knowledge to adjust or evaluate the performance of time series anomaly detection methods can lead to overprediction \cite{darban2025carla}.
We report the standard F1 score in Table \ref{tab:main-table} and the ROC-AUC score \footnote{We attempt to replicate MultiRC results on ROC-AUC, but we only manage to get some positive results for MSL, SMAP, SMD, and PSM. We do not have the optimal hyperparameters for SWaT.} in Table \ref{tab:main-result-auroc}.

\begin{table*}[t]
\centering
\caption{Performance comparison across multiple anomaly detection methods on five datasets: MSL, SMAP, SMD, PSM, and SWAT. The table reports precision (P), recall (R), and F1 scores for each dataset, along with the average F1 score across all datasets. The highest F1 for each dataset is highlighted in bold, while the second-highest is underlined. Methods marked with \(^*\) were replicated in our work, while unmarked methods are sourced from \cite{hu2024multirc}.}

\resizebox{\textwidth}{!}{%

\begin{tabular}{c|ccc|ccc|ccc|ccc|ccc|c}
\hline
Dataset      & \multicolumn{3}{c|}{MSL}       & \multicolumn{3}{c|}{SMAP}      & \multicolumn{3}{c|}{SMD}       & \multicolumn{3}{c|}{PSM}        & \multicolumn{3}{c|}{SWaT}       & Average        \\ \hline
Method       & P     & R     & F1             & P     & R     & F1             & P     & R     & F1             & P     & R      & F1             & P     & R      & F1             & F1             \\ \hline
A.Transformer \cite{xu2022anomaly}        & 14.62 & 12.42 & 13.43          & 11.40 & 22.80 & 15.20          & 10.52 & 19.42 & 13.65          & 30.56 & 34.64  & 32.47          & 75.00 & 60.50  & 66.97          & 28.34          \\
PAD \cite{jhin2023precursor}         & 14.37 & 13.22 & 13.77          & 15.19 & 34.41 & 21.08          & 11.23 & 18.76 & 14.05          & 31.23 & 33.05  & 32.11          & 74.83 & 59.09  & 66.04          & 29.41          \\
% LSTM-NDT     & 16.18 & 15.28 & 15.72          & 11.68 & 22.65 & 15.41          & 14.21 & 16.42 & 15.24          & 39.20 & 40.31  & 39.75          & 66.76 & 68.36  & 67.55          & 30.73          \\
OmniAnomaly \cite{su2019robust}        & 13.66 & 21.66 & 16.75          & 11.90 & 23.39 & 15.77          & 16.22 & 18.19 & 17.15          & 39.45 & 40.88  & 40.15          & 85.73 & 58.57  & 69.59          & 31.88          \\
GANomaly \cite{du2021gan}    & 15.36 & 17.30 & 16.27          & 12.18 & 23.41 & 16.02          & 15.06 & 23.19 & 18.26          & 37.02 & 43.06  & 39.81          & 83.99 & 59.77  & 69.84          & 32.04          \\
CAE-Ensemble \cite{campos2021unsupervised} & 20.35 & 18.27 & 19.25          & 15.88 & 26.68 & 19.91          & 18.41 & 19.51 & 18.94          & 40.18 & 41.99  & 41.07          & 88.61 & 59.90  & 71.48          & 34.13          \\
TimesNet\(^*\) \cite{wu2023timesnet}     & 13.44 & 84.26 & 23.18          & 14.03 & 88.41 & 24.22          & 15.42 & 33.41 & 21.10          & 27.74 & 100.00 & 43.43          & 12.14 & 100.00 & 21.65          & 26.72          \\
D3R \cite{wang2024drift}         & 19.99 & 20.32 & 20.15          & 15.73 & 26.89 & 19.85          & 15.78 & 19.33 & 17.38          & 40.73 & 42.21  & 41.46          & 84.13 & 61.85  & 71.29          & 34.02          \\
% PUAD         & 20.04 & 18.74 & 19.37          & 15.20 & 27.20 & 19.50          & 16.37 & 21.85 & 18.72          & 39.91 & 42.38  & 41.11          & 85.68 & 58.05  & 69.21          & 33.58          \\
IGCL \cite{zhao2024unsupervised}        & 20.43 & 21.69 & 21.04          & 17.09 & 27.09 & 20.96          & 16.44 & 25.65 & 20.04          & 41.22 & 44.99  & 43.02          & 84.21 & 63.19  & 72.20          & 35.45          \\ 
MultiRC \cite{hu2024multirc}     & 14.74 & 69.84 & { \ul 24.34}          & 16.59 & 47.11 & { \ul 24.54}          & 19.99 & 23.27 & { \ul 21.51}          & 36.77 & 65.30  & { \ul 47.05}          & 99.38 & 60.29  & {\ul75.05}          & {\ul 38.50}    \\ \hline
% MultiRC \(^*\) \cite{hu2024multirc}     & 28.77 & 15.00 & { \ul 19.72}          & 10.26 & 4.33 & { \ul 6.09}          & 17.10 & 32.94 & { \ul 22.52}          & 27.74 & 100.00  & { \ul 43.43}          & 6.96 & 15.72  & 9.64          & {\ul 20.28}    \\ \hline
Deep SVDD \cite{ruff2018deep}   & 12.94 & 16.04 & 14.32          & 11.77 & 17.09 & 13.94          & 14.99 & 15.78 & 15.37          & 36.63 & 36.30  & 36.46          & 81.29 & 61.95  & 70.31          & 30.08          \\
DAGMM \cite{zong2018deep}       & 13.94 & 17.04 & 15.33          & 11.91 & 22.00 & 15.45          & 14.74 & 15.37 & 15.05          & 39.27 & 36.14  & 37.64          & 82.14 & 61.72  & 70.48          & 30.79          \\
DCDetector \cite{yang2023dcdetector}  & 2.35  & 3.46  & 2.80           & 8.04  & 11.30 & 9.40           & 3.71  & 9.03  & 5.26           & 28.97 & 12.05  & 17.02          & 46.45 & 36.36  & 40.79          & 15.05          \\ \hline
VISTA (Ours) & 37.59 & 54.52 & \textbf{44.50} & 29.95 & 43.18 & \textbf{35.37} & 36.85 & 24.75 & \textbf{29.61} & 36.64 & 75.83  & \textbf{49.41} & 90.67 & 64.74  & \textbf{75.54}    & \textbf{46.88} \\ \hline
\end{tabular}
}

\label{tab:main-table}
\end{table*}

\begin{table*}[t]
\centering
\caption{Performance comparison across multiple anomaly detection methods on five datasets: MSL, SMAP, SMD, PSM, and SWAT. The table reports the ROC-AUC for each dataset. The best performance for each dataset is highlighted in bold, while the second-best is underlined. 
Methods marked with \(^*\) were replicated in our work, while unmarked methods are sourced from \cite{zhao2024unsupervised}}.

\begin{tabular}{c|ccccc|c}
\hline
Dataset      & MSL            & SMAP           & SMD            & PSM            & SWaT            & Average \\ \hline
A.Transformer \cite{xu2022anomaly}         & 50.21          & 53.70          & 59.30          & 54.91          & 75.63       & 58.75   \\
PAD \cite{jhin2023precursor}         & 48.28          & 50.13          & 49.66          & 51.02          & 61.35         & 52.09 \\
CAE-Ensemble \cite{campos2021unsupervised} & 51.26          & 58.76          & 67.99          & 63.68          & 81.71   & 58.75 \\
TimesNet\(^*\) \cite{wu2023timesnet}    & 60.37          & 44.43          & { \ul 75.69}    & 57.68          & 24.48         & 52.53 \\
D3R \cite{wang2024drift}         & 51.83          &  59.61    & 66.83          & 63.93    & 80.62         & 64.56 \\
IGCL \cite{zhao2024unsupervised}     & 55.19    & {\ul 60.74 }         & 68.75   & {\ul 65.37 }         & { \ul 81.90 }       & { \ul 66.39}  \\ 
MultiRC\(^*\) \cite{hu2024multirc}     & {\ul 76.02}    & 52.45          & \textbf{76.17}    & 57.40          & 22.81        & 56.97  \\ \hline
DCDetector\(^*\) \cite{yang2023dcdetector}  & 30.34          & 43.98          & 50.44          & 50.91          & 65.57       & 48.25   \\
VISTA (Ours) & \textbf{80.86} & \textbf{63.52} & 73.14 & \textbf{66.39} & \textbf{86.98} & \textbf{73.70} \\ \hline
\end{tabular}
\label{tab:main-result-auroc}
\end{table*}

\subsection{Discussion}
\label{sec:discussion}
VISTA outperforms state-of-the-art baselines, achieving the highest average F1 score of 46.88, which represents a substantial improvement over the previous state-of-the-art method, MultiRC (38.50). This performance gain underscores the robustness and versatility of VISTA’s design, particularly in its ability to model complex temporal correlations and mitigate noise through its decomposition and self-attention strategies. VISTA demonstrates significant improvements over current state-of-the-art methods by 21.32\%, 11.15\%, 8.51\%, 6.39\%, and 3.37\% on MSL, SMAP, SMD, PSM, and SWaT respectively. 
These results highlight VISTA’s ability to detect anomalies effectively across diverse datasets.

The ROC-AUC results further demonstrate VISTA’s superiority in multivariate time series anomaly detection. 
VISTA achieved overall best performance compared to previous methods by achieving an average ROC-AUC of 73.70\%.
In comparison, DCDetector and PAD show limited performance, likely due to their inability to handle intricate dependencies or noise. 
Reconstruction-based methods like TimesNet, IGCL, and MultiRC perform well on datasets such as SMD and SWaT, but lack consistency across diverse scenarios. 
Specifically, MultiRC outperforms VISTA by 3.03\% on the SMD dataset but fails to detect anomalies effectively in the SWaT dataset.
Reconstruction methods will perform better in this case because they have more training samples to train their model (approximately 1.4 times the dataset size of SWaT).
Furthermore, each variable in SMD represents complex and diverse patterns of normality, which introduces significant challenges in modeling the memory bank. 
The high variance of normal training samples makes it difficult for VISTA to generalize within a limited memory resource, the memory bank struggles to capture all representative feature vectors that effectively encapsulate such diverse and dynamic behaviors.

On the other hand, the SWaT dataset exhibits more structured and consistent temporal patterns, allowing VISTA to leverage its temporal dependency modeling and feature extraction capabilities to detect anomalies more effectively. 
This highlights a key limitation of VISTA in environments like SMD, where the diversity and complexity of training data require a larger memory bank to model the normal training samples. 
Conversely, reconstruction methods appear better suited to handling the diverse normal behaviors in SMD, possibly due to more training data to train the model. 
However, its inability to handle datasets like SWaT suggests a lack of adaptability in capturing structured temporal dependencies, where VISTA excels.

\begin{figure*}[tp]
    \centering
    \includegraphics[width=\linewidth]{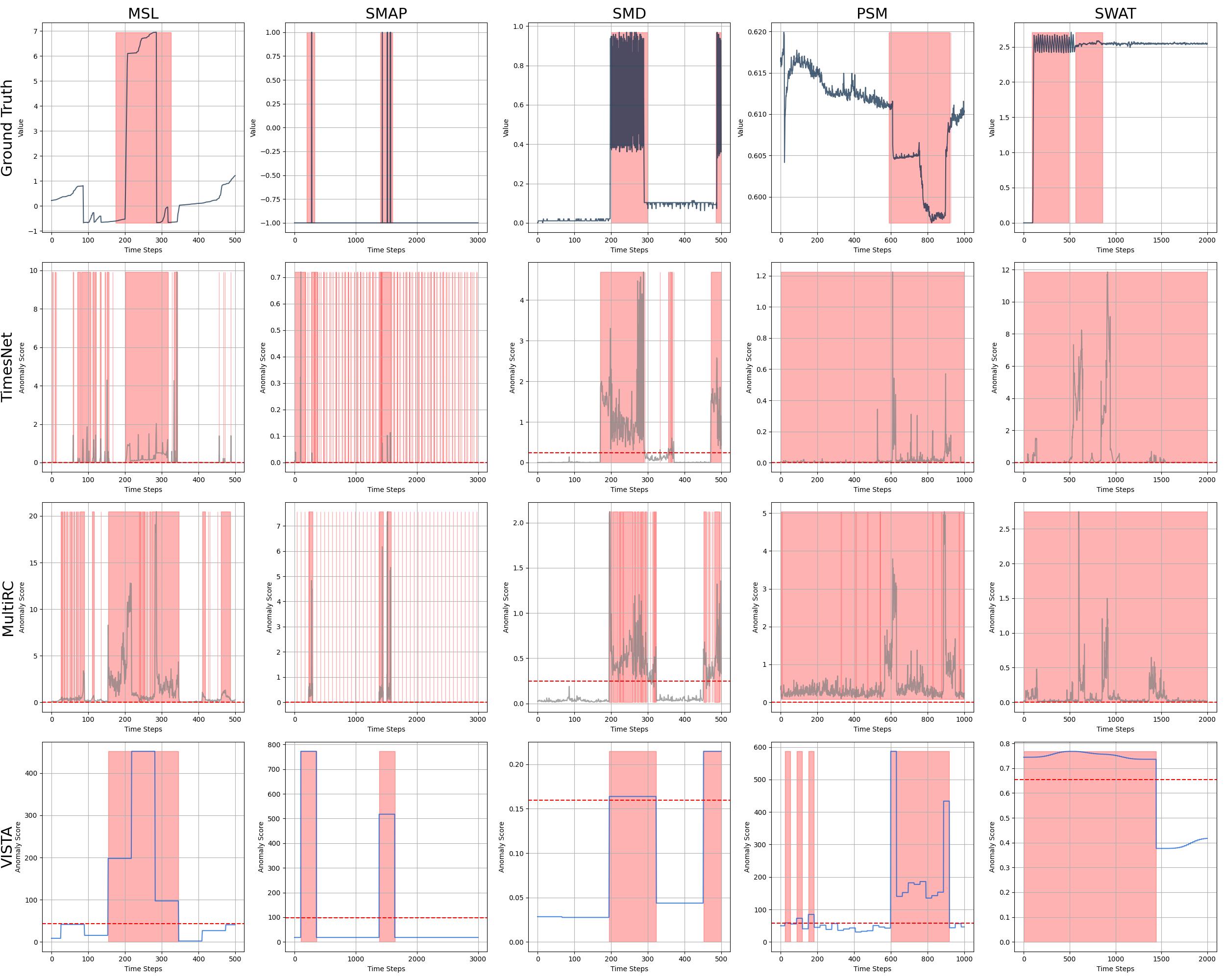}
    \caption{
    Qualitative comparison of anomaly detection results for TimesNet, MultiRC, and VISTA. The first row shows the raw time series data with anomalies highlighted in red. The second row depicts anomaly scores predicted by TimesNet, with the dotted horizontal line representing the threshold calculated by Optimal F1 and predicted anomalies highlighted in red. The third row shows anomaly scores from DCDetector. The final row presents VISTA’s patch-wise predictions, which result in square-like anomaly regions compared to previous methods. VISTA demonstrates the closest alignment with the ground truth.
    }
    \label{fig:qualitative}
\end{figure*}

TimesNet and MultiRC exhibit competitive performance but struggle with noisy data. For example, datasets like SMD and PSM feature significant noise, which results in high false prediction rates for many baseline methods. In contrast, VISTA demonstrates robustness under such conditions. Figure \ref{fig:qualitative} visually contrasts the outputs of TimesNet, MultiRC and VISTA, highlighting VISTA’s superior ability to detect anomalies effectively across various datasets.

\section{Ablation Study}
\label{sec:ablation}

\subsection{2D Temporal Representation}
\begin{figure}[tp]
    \centering
    \includegraphics[width=\linewidth]{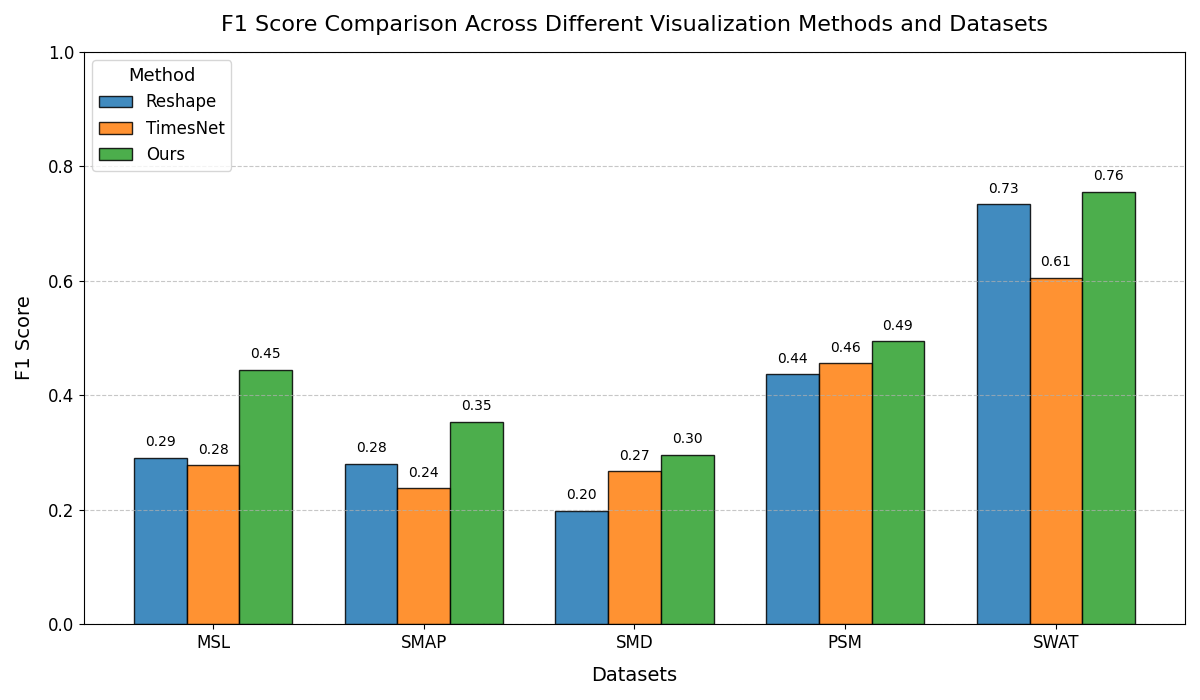}
    \caption{
    Comparison between different 2D temporal representations across datasets. VISTA consistently outperforms other methods in the F1 Score. 
    }
    \label{fig:abla-tcm}
\end{figure}

One of the key contributions of VISTA is its ability to unify time series anomaly detection with image-based anomaly detection. Previous methods fail to provide effective representations of temporal dependency for anomaly detection algorithms. In contrast, the proposed temporal correlation matrix not only offers a clear visualization of time series correlations but also enables efficient feature extraction using a pretrained CNN, underscoring VISTA’s effectiveness in providing a robust visual representation. 
To evaluate the utility of the temporal correlation matrix, we compare it against other 1D-to-2D transformations. First, we consider a simple method that reshapes a time series of length 1024 into a \(32 \times 32\) representation. Next, we replace VISTA’s temporal self-attention mechanism with TimesNet’s 2D representation. Figure \ref{fig:abla-tcm} presents the results of these comparisons. The temporal correlation matrix consistently outperforms alternative methods, highlighting its effectiveness as a visual representation for time series anomaly detection.

% include figure showcasing methods

\subsection{Reliance for Seasonal Period for STL Decomposition}
\label{sec:abla-decompose-ratio}
\begin{figure}[tp]
    \centering
    \includegraphics[width=\linewidth]{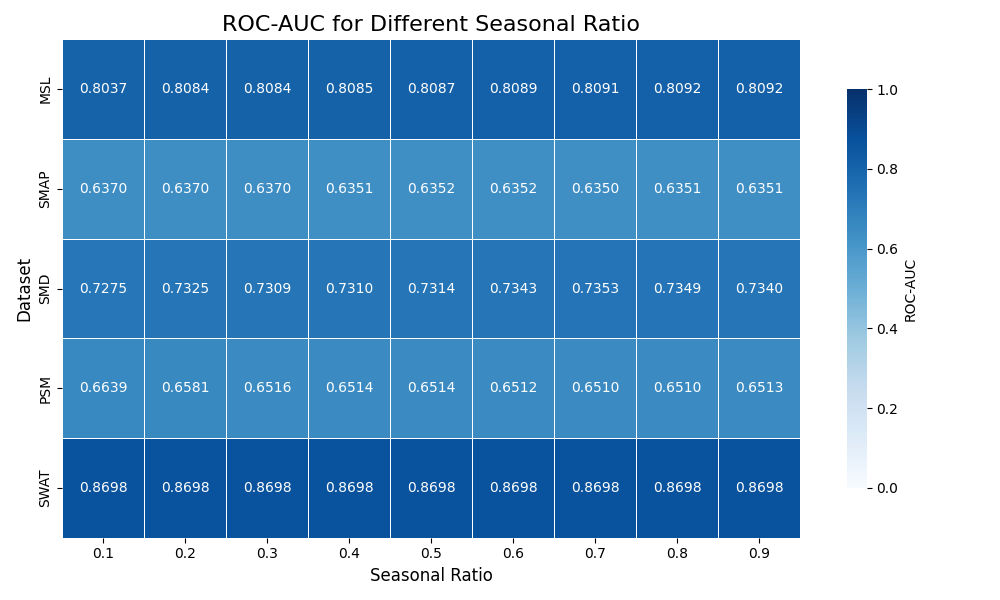}
    \caption{
    Effect for Seasonal Ratio for STL Decomposition on ROC-AUC.
    VISTA consistently shows minimal reliance on seasonal decomposition period for all 5 datasets.
    }
    \label{fig:abla-seasonal}
\end{figure}

In this section, we evaluate the impact of different seasonal decomposition ratios on the AUROC of our proposed method across all five datasets. 
We define the seasonal decomposition ratio as the ratio between seasonal period and the window size.
We vary the seasonal decomposition ratio from 0.1 to 0.9 to observe the reliance of VISTA on STL decomposition for each window.
Figure \ref{fig:abla-seasonal} shows that VISTA maintains stable performance across different seasonal decomposition ratios for all datasets. 

% Use figure* for multi-column figure

 This ablation study demonstrates that VISTA’s performance is not overly reliant on specific seasonal configurations, ensuring reliability and adaptability across diverse datasets and conditions.

\subsection{Decomposition Component Selection}
\label{sec:abla-decompose-component}
This section explores the impact of using different decomposition components (trend, seasonal, and residual) on anomaly detection performance across five datasets. Additionally, we compare these decompositions against the use of the original time series data for temporal correlation matrix computation. The results of this comparison are summarized in Table \ref{tab:abla-decomposition}.

\begin{table}[t]
\centering
\caption{Analysis of the impact of different decomposition types across all five datasets. 
The table shows the F1-score for different combinations of the original time series and its decomposed components (trend, seasonal, residual) for the construction of the temporal correlation matrix. 
The best performance for each dataset is highlighted in bold, and the second-best performance is underlined.}

\resizebox{\columnwidth}{!}{

\begin{tabular}{cccc|ccccc}
\hline

\multicolumn{4}{c|}{Decomposition Type} & \multicolumn{5}{c}{Dataset}                                                                  \\ \hline
Original  & Trend & Seasonal & Residual & MSL              & SMAP             & SMD              & PSM              & SWAT             \\ \hline
\cmark & \xmark& \xmark& \xmark& 0.3895          & 0.2914          & {\ul 0.2989}    & 0.4574          & 0.6849                \\
\xmark& \cmark & \xmark& \xmark& {\ul 0.4064}    & {\ul 0.2952}    & \textbf{0.3025} & {\ul 0.4581}    & {\ul 0.6850}          \\
\xmark& \xmark& \cmark & \xmark& 0.2866          & 0.2271          & 0.0798          & 0.4345          & 0.2171                \\
\xmark& \xmark& \xmark& \cmark & 0.3199          & 0.2271          & 0.2207          & 0.4345          & 0.2171                \\
\xmark& \cmark & \cmark & \cmark & \textbf{0.4450} & \textbf{0.3537} & 0.2961          & \textbf{0.4941} & \textbf{0.7554} \\ \hline
\end{tabular}

}

\label{tab:abla-decomposition}

\end{table}

Leveraging only the trend component demonstrates improved results compared to using the original time series data, as irregularities in the raw input signal are mitigated through smoothing. This highlights the significant impact of noise in time series on anomaly detection. However, such simple denoising can result in the loss of a considerable amount of useful information, as other components also play critical roles in TSAD. For instance, in datasets like MSL and SMD, utilizing the residual component alone still achieves suboptimal yet meaningful anomaly detection performance, underscoring its relevance in identifying anomalies that deviate from modeled trends and seasonal patterns.

To achieve optimal results, we leveraging all decomposed components within VISTA. This approach ensures that the temporal correlation matrix captures both large-scale patterns and fine-grained irregularities, enabling a more comprehensive and robust representation for TSAD.

\subsection{Feature Selection}
\label{sec:abla-model-featuere}
The selection of layers (\text{Layer2}, \text{Layer3}, \text{Layer4}) reflects the utilization of low-level, mid-level, and high-level features for anomaly detection, each contributing distinct information to the process. Low-level features (\text{Layer2}) capture fine-grained patterns and texture details, making them crucial for detecting localized anomalies. However, their sensitivity to noise and inability to model long-term temporal dependencies limit their standalone effectiveness, necessitating integration with higher-level features for more robust detection. Mid-level features (\text{Layer3}) provide abstract representations and intermediate spatial-temporal dependencies, offering a balance between granularity and context. High-level features (\text{Layer4}) encapsulate global representations and semantics, making them indispensable for detecting anomalies characterized by overarching temporal or contextual shifts. These features are particularly effective in datasets where broad patterns dominate. Table \ref{tab:model-feature} reports the F1-scores achieved with different layer selections.

\begin{table}[t]
\centering
\caption{Results for VISTA across different layer combinations. The table shows the F1-Score for all five datasets. The best performance for each dataset is highlighted in bold, and the second-best performance is underlined. The ablation evaluates the contributions of different layer combinations to anomaly detection performance.}

\resizebox{\columnwidth}{!}{

\begin{tabular}{ccc|ccccc}
\hline
\multicolumn{3}{c|}{Feature Layer} & \multicolumn{5}{c}{Dataset}                                                             \\ \hline
Layer2     & Layer3    & Layer4    & MSL             & SMAP            & SMD             & PSM             & SWAT            \\ \hline
\cmark & \xmark& \xmark& 0.4073          & 0.2503          & 0.2924          & 0.4734          & 0.3307          \\
\xmark& \cmark & \xmark& \textbf{0.4542} & 0.2332          & \textbf{0.2999} & 0.4842          & {\ul 0.5324}    \\
\xmark& \xmark& \cmark & 0.4093          & 0.2271          & 0.2932          & 0.5054          & \textbf{0.7554} \\
\cmark & \cmark & \xmark& 0.4180          & {\ul 0.2577}    & 0.2956          & 0.4819          & 0.4578          \\
\cmark & \xmark& \cmark & 0.4082          & 0.2498          & 0.2927          & {\ul 0.4898}    & \textbf{0.7554} \\
\xmark& \cmark & \cmark & {\ul 0.4450}    & \textbf{0.3537} & {\ul 0.2961}    & \textbf{0.4941} & \textbf{0.7554} \\ \hline
\end{tabular}
}

\label{tab:model-feature}

\end{table}

Overall, high-level features yield better performance than low-level features for anomaly detection, as they provide the deeper temporal representations necessary to encode semantic meaning. However, high-level features alone may lack the precision needed to effectively locate anomalies, underscoring the importance of incorporating low-level features for accurate defect detection. To achieve a balance between contextual understanding and localization accuracy, we adopt \text{Layer3} and \text{Layer4} as our default configuration.

\section{Conclusion}
\label{sec-conclusion}
% what we do and what we achieved?
VISTA integrates dynamic time series decomposition, temporal self-attention, and multivariate feature aggregation to improve anomaly detection in multivariate time series data. VISTA demonstrates its robustness and effectiveness in managing noisy and complex temporal environments by achieving the overall best F1 and ROC-AUC performance across five commonly used datasets. Its training-free nature further ensures practicality and computational efficiency, removing the overhead associated with model fine-tuning.

% what is our main works about
% remove noise from raw dataset but still maintain its contribution in TCM
% able to detect different types of anomalies in trend, seasonal, and residual ...
% a new method that bridge image and time series, with the following advantages
This study emphasizes decomposing raw time series into trend, seasonal, and residual components, preserving noise in the residual to capture anomalies, which is crucial for TSAD. 
Moreover, it seamlessly bridges the gap between TSAD and image anomaly detection, by introducing a non-trainable temporal self-attention module that not only captures intricate temporal dependencies but also enhances feature distinctiveness to effectively identify temporal discrepancies.

% future research
Future research directions should focus on developing more sophisticated decomposition methods, which can better preserve residual information while simultaneously enhancing temporal correlation matrices. 
While convolutional neural networks (CNNs) are effective feature extractors, they are inherently limited when applied to temporal correlation matrices, as they are primarily trained on natural images. 
We suggest that future work may explore domain adaptation techniques for better visual representation of temporal dependencies and optimizing feature representation to mitigate memory constraints.

% closing remarks
Our findings, evaluated using reliable metrics such as the F1 score (without point adjustment) and ROC-AUC, underline the need for standardized evaluation practices in TSAD research. To advance the field, we advocate for the consistent use of F1 score and ROC-AUC as standard metrics in future research to ensure robust and meaningful comparisons across methods.

\section{Acknowledgment}
This work was partly supported by the National Key R\&D Program of China (2023YFB430220).

% Please replace this by building a bib file and using bibtex
\bibliographystyle{elsarticle-num}    
\bibliography{ref}

\end{document}